\pdfoutput=1
\documentclass[sigplan,screen,nonacm]{acmart}
\AtBeginDocument{%
  \providecommand\BibTeX{{%
    \normalfont B\kern-0.5em{\scshape i\kern-0.25em b}\kern-0.8em\TeX}}}

\usepackage{tabularx}
\usepackage{multirow}
\usepackage{array}
\usepackage{pgfplots}
\pgfplotsset{width=0.45\textwidth, height=0.35\textwidth, compat=1.7}

\begin{document}

\title{Performance, Transparency and Time. Feature selection to speed up the diagnosis of Parkinson’s disease}
\author{Pierluigi Costanzo}
\email{costanzo.p@live.unic.ac.cy}
\orcid{0000-0002-5000-4881}
\affiliation{
  \institution{University of Nicosia}
  \country{Cyprus}
}

\author{Kalia Orphanou}
\email{orphanou.k@unic.ac.cy}
\affiliation{
  \institution{University of Nicosia}
  \country{Cyprus}
}

\renewcommand{\shortauthors}{Costanzo and Orphanou.}

\begin{abstract}
Accurate and early prediction of a disease allows to plan and improve a patient's quality of future life. During pandemic situations, the medical decision becomes a speed challenge in which physicians have to act fast to diagnose and predict the risk of the severity of the disease, moreover this is also of high priority for neurodegenerative diseases like Parkinson’s disease. Machine Learning (ML) models with Features Selection (FS) techniques can be applied to help physicians to quickly diagnose a disease. FS optimally subset features that improve a model performance and help reduce the number of needed tests for a patient and hence speeding up the diagnosis. This study shows the result of three Feature Selection (FS) techniques pre-applied to a classifier algorithm, Logistic Regression, on non-invasive test results data. The three FS are Analysis of Variance (ANOVA) as filter based method, Least Absolute Shrinkage and Selection Operator (LASSO) as embedded method and Sequential Feature Selection (SFS) as wrapper method. The outcome shows that FS technique can help to build an efficient and effective classifier, hence improving the performance of the classifier while reducing the computation time.
\end{abstract}


\maketitle

\section{Introduction}
Parkinson’s disease (PD) is a neurodegenerative disorder that leads to partial or full loss in motor reflexes, difficulties in speech, mental processing, and other vital functions \cite{doc:61c9e0678f080835edec03ca}. As a degenerative disorder, it gets worse over time. There is no known cure at the moment but only therapies that can help reduce the effects of the difficulties this disease causes\cite{doc:61cc1d7c8f08bf9aa010cb4d}, hence it’s crucial to have tools that can diagnose it at an early stage possible. Voice problem is an effect of PD and tests based on speech data analysis are one of the non-invasive diagnostic tools to diagnose the disease, they are also low cost and easy to self-use \cite{doc:61c9dfe68f08bf9aa01081f7}.

Machine Learning (ML) techniques applied to biomedical data\cite{doc:61c9e1b88f083801efa8c721} have to fulfill critical characteristics of the medical decision process\cite{doc:61c9e1ce8f0808703bd4405b}: performance, transparency, time. It is vital that a patient diagnosis is correct and, hence the ML algorithm predicts the diagnosis as accurately as possible. Trust is a crucial component between the physician and the patient for the physician to give and for the patient to get the best possible care. This is translated in ML as model explainability, so called “glass-box” ML algorithms are models whose output is explainable and they can be used trustworthy in the process of medical decisions. Moreover, collecting tests for a patient diagnosis is expensive and time-consuming, and in critical situations, time might not be an option.

\vspace{2 mm}
This research study made use of data from non-invasive tests on healthy and PD patients. As the dataset include multiple record for each individual, stratified leave-one-subject-out (LOSO) cross-validation technique was used. Three FS methods were tested with two different metric optimization  techniques, accuracy and cross-entropy. Different models were output and the results were compared with each other and with a baseline model without any FS technique. The evaluation was intended as classifying individual by the majority of the multiple records for each of them.

\vspace{2 mm}
In Section 2 the dataset used for the early diagnosis of Parkinson is described. In Section 3 the methodology to test the different FS techniques is described, which ML classifier has been used, the data preprocessing steps, more details about the three FS, evaluation of the best feature selected and evaluation for comparing the results of the different models. Section 4 is an overview of the results and Conclusions are in Section 5.

\section{Dataset}
\begin{table*}
  \caption{Features Summary}
  \label{tab:features summary}
  \begin{tabularx}{\textwidth} {
        >{\centering\arraybackslash}p{0.01\textwidth}
        >{\centering\arraybackslash}p{0.28\textwidth}
        >{\centering\arraybackslash}p{0.24\textwidth}
        >{\centering\arraybackslash}p{0.07\textwidth}
        >{\centering\arraybackslash}p{0.06\textwidth}
        >{\centering\arraybackslash}p{0.08\textwidth}
        >{\centering\arraybackslash}p{0.05\textwidth}
        >{\centering\arraybackslash}p{0.05\textwidth}
    }
    \toprule
    ID & Feature & Description & Type & MEAN & STD. DEV & MIN & MAX\\
    \midrule
    1 & Jitter (local) & \multirow{5}{=}{\centering Variation in the frequency of the sound} & FLOAT & 2.68 & 1.78 & .19 & 14.38\\
    2 & Jitter (local, absolute) & & FLOAT & .0002 & .0001 & .000006 & .0008\\
    3 & Jitter (rap) & & FLOAT & 1.25 & .98 & .06 & 8.02\\
    4 & Jitter (ppq5) & & FLOAT & 1.35 & 1.14 & .08 & 13.54\\
    5 & Jitter (ddp) & & FLOAT & 3.74 & 2.94 & .19 & 24.05\\
    \midrule
    6 & Shimmer (local) & \multirow{6}{=}{\centering Variation in the amplitude of the sound} & FLOAT & 12.92 & 5.45 & 1.19 & 41.14\\
    7 & Shimmer (local, dB) & & FLOAT & 1.19 & .42 & .10 & 2.72\\
    8 & Shimmer (apq3) & & FLOAT & 5.70 & 3.02 & .50 & 25.82\\
    9 & Shimmer (apq5) & & FLOAT & 7.98 & 4.84 & .71 & 72.86\\
    10 & Shimmer (apq11) & & FLOAT & 12.21 & 6.02 & .52 & 44.76\\
    11 & Shimmer (dda) & & FLOAT & 17.10 & 9.05 & 1.49 & 77.46\\
    \midrule
    12 & Autocorrelation & \multirow{3}{=}{\centering Noise in the sound assessment} & FLOAT & .85 & .09 & .54 & .99\\
    13 & Noise-to-Harmonic & & FLOAT & .23 & .15 & .002 & .87\\
    14 & Harmonic-to-Noise & & FLOAT & 9.99 & 4.29 & .70 & 28.42\\
    \midrule
    15 & Median pitch & \multirow{5}{=}{\centering How high or low is the sound based on the frequency of vibration of the sound waves produced} & FLOAT & 163.37 & 56.02 & 81.46 & 468.62\\
    16 & Mean pitch & & FLOAT & 168.73 & 55.97 & 82.36 & 470.46\\
    17 & Standard dev. of pitch & & FLOAT & 27.55 & 36.67 & 0.53 & 293.88\\
    18 & Minimum pitch & & FLOAT & 134.54 & 47.06 & 67.96 & 452.08\\
    19 & Maximum pitch & & FLOAT & 234.86 & 121.54 & 85.54 & 597.97\\
    \midrule
    20 & Number of pulses & \multirow{4}{=}{\centering Pulse related metrics} & INT & 109.74 & 150.03 & 0 & 1490\\
    21 & Number of periods & & INT & 105.97 & 149.42 & 0 & 1489\\
    22 & Mean period & & FLOAT & .007 & .002 & .002 & .01\\
    23 & Standard dev. of period & & FLOAT & .001 & .001 & .0001 & .01\\
    \midrule
    24 & Fraction of unvoiced frames & \multirow{3}{=}{\centering Voice related metrics} & FLOAT & 27.68 & 20.98 & 0.00 & 88.16\\
    25 & Number of voice breaks & & INT & 1.13 & 1.61 & 0 & 12\\
    26 & Degree of voice breaks column & & FLOAT & 12.37 & 15.16 & 0.00 & 69.12\\
    \bottomrule
  \end{tabularx}
\end{table*}

The dataset used is the result of a study made on 40 individuals, 20 People with Parkinsonism (PWP) and 20 healthy individuals, who voluntarily participated in the research at the Department of Neurology in Cerrahpaşa Faculty of Medicine, Istanbul University \cite{doc:61b768d98f08073661ec2e97}. All individuals recorded 26 voice samples of sustained vowels, numbers, words, and short sentences. The Table \ref{tab:features summary} summarizes the features in the dataset.

There are 5 statistics of Pitch of vocal oscillation, which measure how high or low is the sound based on the frequency of vibration of the sound waves produced: median, mean, standard deviation, minimum, maximum.
There are 5 Jitter metrics, which measure the variation in the frequency of the sound: jitter in \%, absolute jitter in microseconds, jitter as relative amplitude perturbation (rap), jitter as 5-point period perturbation quotient (ppq5) and jitter as average absolute difference of differences between jitter cycles (ddp).
There are 6 Shimmer metrics, which measure the variation in the amplitude of the sound: shimmer in \%, absolute shimmer in decibels (dB), shimmer as 3-point, 5-point and 11-point amplitude perturbation quotient (apq3, apq5, apq11), and shimmer as average absolute differences between the amplitudes of shimmer cycles (dda).
There are Noise-to-Harmonic (NHR) and Harmonic-to-Noise (HNR) ratios that assess the noise in the sound, and a correlation of the two.
There are 4 pulse related metrics: number of pulses, number of periods mean of period and standard deviation of period.
There are 3 voice related metrics: fraction of unvoiced frames, number and degree of voice breaks.

The dataset also includes the binary class information, which is used for training and testing the model, and the Unified Parkinson’s Disease Rating Scale (UPDRS) score which instead has been excluded in this study.

\section{Methodology}
\subsection{Proposed Method}
The dataset is composed of 40 individuals and 26 recordings for each individual. A stratified leave-one-subject-out (LOSO)\cite{RefWorks:RefID:49-sakar2010telediagnosis} k-fold (with k=4) cross-validation technique was used to divide the training dataset for evaluating the performance of each FS method. As the number of subjects is only 40, k=4 was used to evaluate the model 4 times with 25\% of the subjects held out as validation set. The LOSO was used to avoid data leakage, hence all the recordings of a subject were or in the held in training set or in the held out validation set. The stratification was used to balance PWP and healthy individuals in the training and validation set.

At each iteration, a preprocessing step was applied (power transformation and standardization) and three feature selection techniques were tested with two different metrics optimization, maximization of the accuracy and minimization of the cross-entropy loss.

The Logistic Regression model was the algorithm chosen. It is a linear model, a gradient descent based algorithm used in classification problems where the probabilities of the outcome are modeled with a logistic function. A binary logistic model, as in this study, has a dependent variable with only two possible outcomes. Moreover as stated in Section 1, model explainability is important, hence the choice of this model as Logistic Regression is a “glass-box” ML algorithm and the output is easily explainable. 

The performance of the Logistic Regression model trained with all the features was compared with each performance of the model trained with the best subset of features from each FS techniques tested.

\subsection{Preprocessing the data}
In Machine Learning, preprocessing the data is important for training and testing effectively and efficiently a gradient descent based algorithms classifier, like logistic regression. The difference in feature ranges, variance and skewness, can influence the algorithm and make it difficult and slower to converge toward the minima (error minimization). In this study, a Power Transformation (Yeo-Johnson transformation \cite{doc:61cc4e8e8f0808703bd49825}) and a Standardization step was applied. At each iteration of the k-fold, the transformation steps were fit with the train fold and applied to both train and test fold before training and evaluating the model.

\subsection{Feature selection techniques}
The aim of feature selection techniques is to discard redundant features from the dataset before training a ML model, improving the performance of the model and reducing the execution time. In this study, three feature selection techniques were tested, Analysis of Variance (ANOVA) as filter based method, Least Absolute Shrinkage and Selection Operator (LASSO) as embedded method and Sequential Feature Selection with forward selection (SFS) as wrapper method.

\vspace{2 mm}
Filter based FS methods rank the features based on their independence of the class labels \cite{doc:61c9e5a38f08dfade47e620e}. ANOVA and $\chi^2$ tests calculate the dependency between two variables, the first is used between numerical and categorical variables, the second for categorical variables only. In this study the ANOVA test was used between all the input features and the class as the dataset contains numerical features. The advantage of using these techniques is that they are light computationally and there is less risk of overfitting, on the other hand the features selected might not be an optimal choice as redundant features are also captured as the test is independently performed between each feature and the class. From the 1st to all features ranked, the technique was evaluated against accuracy and cross-entropy loss scores to select the best performing features.

\vspace{2 mm}
Wrapper based FS methods are greedy search approaches which perform an evaluation of the best possible subset of feature combination against an evaluation criteria \cite{doc:61c9e5a38f08dfade47e620e}. There are Selection Algorithms and Heuristic Search Algorithms and for this study the SFS was tested, evaluating the subset of features against accuracy and cross-entropy loss scores. These techniques are computationally expensive and can lead to overfitting.

\vspace{2 mm}
Embedded based FS methods include FS during the training process and they are techniques which might vary from ML algorithm to algorithm\cite{doc:61c9e5a38f08dfade47e620e}. In this study the LASSO (L1 regularization) \cite{doc:61c9fba88f0808703bd441b5} technique was tested and evaluated against accuracy and cross-entropy loss scores.

\vspace{2 mm}
The accuracy score measures the percentage of correctly classified class. In this study is intended as correctly classified individuals by the majority of the 26 records classified for each individual. For the FS process, the subset of features was chosen by maximizing the accuracy score.

The cross-entropy loss score, or negative log-likelihood, measure the distance between the predicted class probability and the actual class. In this study, it is intended as the distance between the predicted classified individuals probability, the mean of the predicted class probability of the 26 records classified for each individual, and the actual classified individuals. For the FS process, the subset of features was chosen by minimizing the cross-entropy loss score (see Eq.~\ref{eq:cross_entropy}).
\begin{equation} \label{eq:cross_entropy}
  H(p,q) = - \sum_{i}p_{i}\log{q_{i}}
\end{equation}
with p the actual class label, which in binary classification is 0 and 1, and q the predicted class probability.

\begin{table*}
  \caption{Results Summary}
  \label{tab:results summary}
  \begin{tabularx}{\textwidth} {
        >{\centering\arraybackslash}p{0.09\textwidth}
        >{\centering\arraybackslash}p{0.15\textwidth}
        >{\centering\arraybackslash}p{0.07\textwidth}
        >{\centering\arraybackslash}p{0.09\textwidth}
        >{\centering\arraybackslash}p{0.08\textwidth}
        >{\centering\arraybackslash}p{0.08\textwidth}
        >{\centering\arraybackslash}p{0.08\textwidth}
        >{\centering\arraybackslash}p{0.09\textwidth}
        >{\centering\arraybackslash}p{0.09\textwidth}
    }
    \toprule
    FS & Strategy & \# FS & 
    Accuracy & Specificity & Sensitivity (Recall)  & 
    Precision & F1-Score & MCC\\
    \midrule
    - & - & 26 & 
    .525 (.150) & .550 (.191) & .525 (.150) & 
    .527 (.162) & .514 (.153) & .052 (.311)\\
    \midrule
    \multirow{2}{=}{\centering ANOVA} & MAX accuracy & 1 &
    .675 (.096) & .650 (.342) & .675 (.096) & 
    .736 (.109) & .653 (.116) & .399 (.191)\\
    & MIN cross-entropy & 7 & 
    .600 (.245) & .600 (.283) & .600 (.245) & 
    .607 (.255) & .594 (.248) & .207 (.500)\\
    \midrule
    \multirow{2}{=}{\centering SFS} & MAX accuracy & 1 &
    .675 (.096) & .650 (.342) & .675 (.096) & 
    .736 (.109) & .653 (.116) & .399 (.191)\\
    & MIN cross-entropy & 3 &
    \textbf{.700} (.082) & .650 (.300) & .700 (.082) & 
    .749 (.107) & \textbf{.686} (.086) & \textbf{.445} (.182)\\
    \midrule
    \multirow{2}{=}{\centering LASSO} & MAX accuracy & 4 &
    .625 (.171) & .650 (.300) & .625 (.171) & 
    .672 (.209) & .611 (.165) & .293 (.375)\\
    & MIN cross-entropy & 9 & 
    .575 (.171) & .600 (.283) & .575 (.171) & 
    .594 (.197) & .567 (.167) & .168 (.367)\\
    \bottomrule
  \end{tabularx}
\end{table*}

\subsection{Model + Feature selected performance evaluation}
For each FS technique, the best features were selected based on maximization of the accuracy and minimization of the cross-entropy loss. The filter and embedded based FS methods are ranking methods, hence the subsets of features were extracted by ranking them and comparing the accuracy and cross-entropy loss for 1 to 25 features. The wrapper based FS methods are not ranking methods hence the subsets of features were extracted by comparing the accuracy and cross-entropy loss for the best subsets of 1 to 25 features.

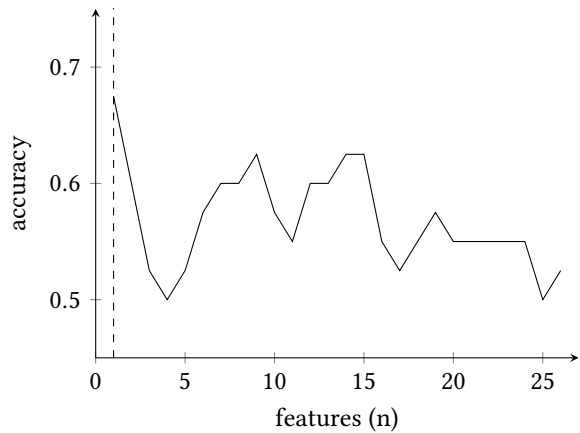
\begin{figure}
\begin{tikzpicture}
\begin{axis}[
axis lines = left, xlabel={features (n)}, ylabel = {accuracy},
xmin=0, xmax=27, ymin=.45, ymax=.75, ymajorgrids=false, grid style=dashed,
]
\addplot[black]
    coordinates {
(1,0.675)(2,0.6)(3,0.525)(4,0.5)(5,0.525)(6,0.575)(7,0.6)
(8,0.6)(9,0.625)(10,0.575)(11,0.55)(12,0.6)(13,0.6)(14,0.625)
(15,0.625)(16,0.55)(17,0.525)(18,0.55)(19,0.575)(20,0.55)(21,0.55)
(22,0.55)(23,0.55)(24,0.55)(25,0.5)(26,0.525)};
\addplot [black, dashed] coordinates {(1, 0) (1, 1)};
\end{axis}
\end{tikzpicture}
\caption{The Accuracy performance of the ranked features with the ANOVA technique. The dashed line indicate the best performance (highest accuracy).}
\label{fig:1}
\end{figure}

\begin{figure}
\begin{tikzpicture}
\begin{axis}[
axis lines = left, xlabel={features (n)}, ylabel = {cross-entropy},
xmin=0, xmax=27, ymin=.5, ymax=.75, ymajorgrids=false, grid style=dashed,
]
\addplot[black]
    coordinates {
(1,0.672)(2,0.69)(3,0.702)(4,0.695)(5,0.691)(6,0.662)(7,0.648)
(8,0.652)(9,0.659)(10,0.657)(11,0.652)(12,0.658)(13,0.659)(14,0.658)
(15,0.658)(16,0.661)(17,0.674)(18,0.672)(19,0.67)(20,0.673)(21,0.676)
(22,0.678)(23,0.688)(24,0.688)(25,0.697)(26,0.696)};
\addplot [black, dashed] coordinates {(7, 0) (7, 1)};
\end{axis}
\end{tikzpicture}
\caption{The cross-entropy performance of the ranked features with the ANOVA technique. The dashed line indicate the best performance (lowest cross-entropy).}
\label{fig:2}
\end{figure}
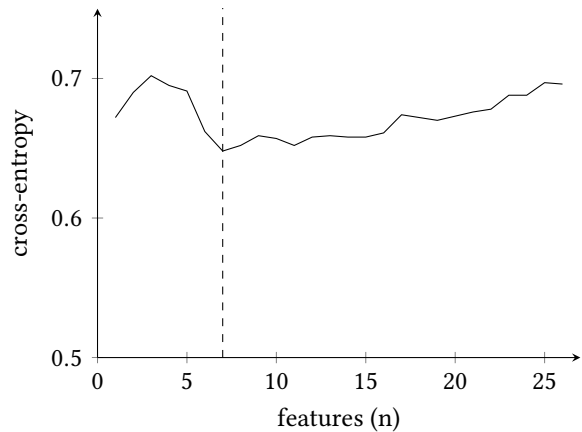

\vspace{2 mm}
Figure \ref{fig:1} and Figure \ref{fig:2} describe the performance of the FS filter technique. For each combination of features, from 1 to 25, the performance of the accuracy and the cross-entropy has been evaluated. Based on the accuracy (figure \ref{fig:1}), 1 feature yield the best metric. Based on the cross-entropy (figure \ref{fig:2}), 7 features yield the best metric.

\begin{figure}
\begin{tikzpicture}
\begin{axis}[
axis lines = left, xlabel={features (n)}, ylabel = {accuracy},
xmin=0, xmax=27, ymin=.45, ymax=.75, ymajorgrids=false, grid style=dashed,
]
\addplot[black]
    coordinates {
(1,0.6)(2,0.55)(3,0.55)(4,0.625)(5,0.55)(6,0.525)(7,0.575)
(8,0.575)(9,0.575)(10,0.575)(11,0.6)(12,0.575)(13,0.575)(14,0.6)
(15,0.55)(16,0.525)(17,0.525)(18,0.55)(19,0.525)(20,0.575)(21,0.575)
(22,0.55)(23,0.525)(24,0.55)(25,0.525)(26,0.525)};
\addplot [black, dashed] coordinates {(4, 0) (4, 1)};
\end{axis}
\end{tikzpicture}
\caption{The Accuracy performance of the ranked features with the LASSO technique. The dashed line indicate the best performance (highest accuracy).}
\label{fig:3}
\end{figure}
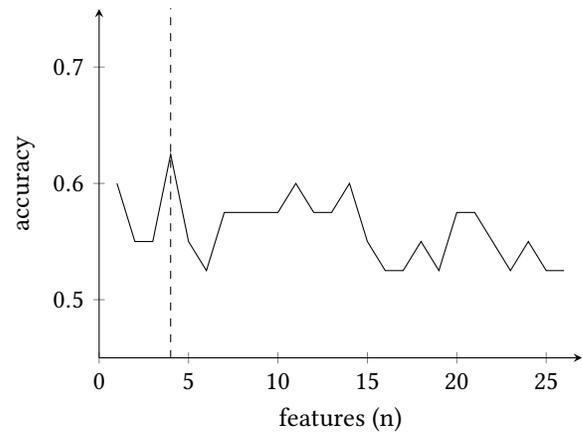

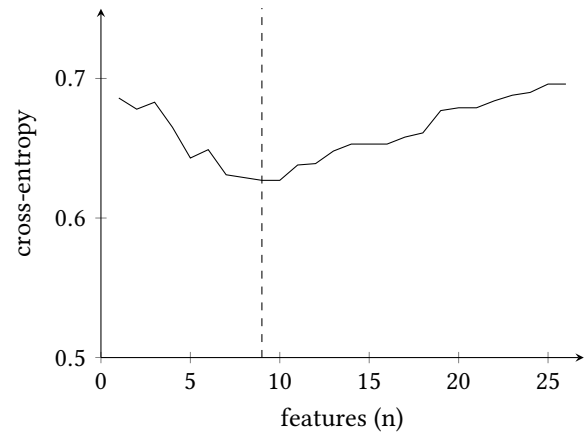
\begin{figure}
\begin{tikzpicture}
\begin{axis}[
axis lines = left, xlabel={features (n)}, ylabel = {cross-entropy},
xmin=0, xmax=27, ymin=.5, ymax=.75, ymajorgrids=false, grid style=dashed,
]
\addplot[black]
    coordinates {
(1,0.686)(2,0.678)(3,0.683)(4,0.665)(5,0.643)(6,0.649)(7,0.631)
(8,0.629)(9,0.627)(10,0.627)(11,0.638)(12,0.639)(13,0.648)(14,0.653)
(15,0.653)(16,0.653)(17,0.658)(18,0.661)(19,0.677)(20,0.679)(21,0.679)
(22,0.684)(23,0.688)(24,0.69)(25,0.696)(26,0.696)};
\addplot [black, dashed] coordinates {(9, 0) (9, 1)};
\end{axis}
\end{tikzpicture}
\caption{The cross-entropy performance of the ranked features with the LASSO technique. The dashed line indicate the best performance (lowest cross-entropy).}
\label{fig:4}
\end{figure}

\vspace{2 mm}
Similarly to the FS embedded technique, for each combination of features, ranked, the performance of the accuracy and the cross-entropy has been evaluated. Based on the accuracy (figure \ref{fig:3}), 4 features yield the best metric. Based on the cross-entropy (figure \ref{fig:4}), 9 features yield the best metric.

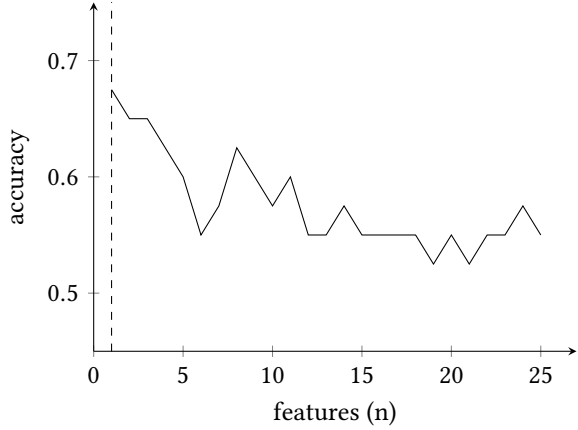
\begin{figure}
\begin{tikzpicture}
\begin{axis}[
axis lines = left, xlabel={features (n)}, ylabel = {accuracy},
xmin=0, xmax=27, ymin=.45, ymax=.75, ymajorgrids=false, grid style=dashed,
]
\addplot[black]
    coordinates {
(1,0.675)(2,0.65)(3,0.65)(4,0.625)(5,0.6)(6,0.55)(7,0.575)
(8,0.625)(9,0.6)(10,0.575)(11,0.6)(12,0.55)(13,0.55)(14,0.575)
(15,0.55)(16,0.55)(17,0.55)(18,0.55)(19,0.525)(20,0.55)(21,0.525)
(22,0.55)(23,0.55)(24,0.575)(25,0.55)};
\addplot [black, dashed] coordinates {(1, 0) (1, 1)};
\end{axis}
\end{tikzpicture}
\caption{The accuracy performance with the accuracy strategy of the subset of of 1 to 25 features with the SFS technique. The dashed line indicate the best performance (highest accuracy).}
\label{fig:5}
\end{figure}

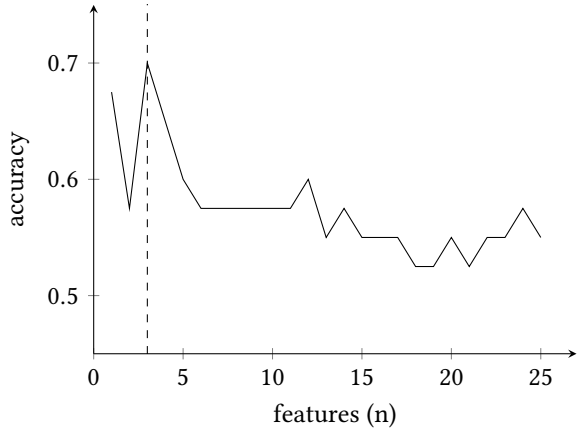
\begin{figure}
\begin{tikzpicture}
\begin{axis}[
axis lines = left, xlabel={features (n)}, ylabel = {accuracy},
xmin=0, xmax=27, ymin=.45, ymax=.75, ymajorgrids=false, grid style=dashed,
]
\addplot[black]
    coordinates {
(1,0.675)(2,0.575)(3,0.7)(4,0.65)(5,0.6)(6,0.575)(7,0.575)
(8,0.575)(9,0.575)(10,0.575)(11,0.575)(12,0.6)(13,0.55)(14,0.575)
(15,0.55)(16,0.55)(17,0.55)(18,0.525)(19,0.525)(20,0.55)(21,0.525)
(22,0.55)(23,0.55)(24,0.575)(25,0.55)};
\addplot [black, dashed] coordinates {(3, 0) (3, 1)};
\end{axis}
\end{tikzpicture}
\caption{The accuracy performance with the cross-entropy strategy of the subset of of 1 to 25 features with the SFS technique. The dashed line indicate the best performance (highest accuracy).}
\label{fig:6}
\end{figure}

\vspace{2 mm}
Differently to the FS filter and embedded technique, for the wrapper based FS method, the subsets of features were extracted by comparing the metrics for the best subsets of 1 to 25 features. In other words, the SFS methodology starts from an empty subset of features, and each feature is added to the subset if yield better result in terms of metric strategy. Figure \ref{fig:5} and Figure \ref{fig:6} describe the performance of the strategy of maximization of the accuracy and minimization of the cross-entropy loss respectively, in terms of accuracy performance of the subset selected. Based on the maximization of the accuracy strategy (Figure \ref{fig:5}), 1 feature yields the best performance. Based on the minimization of the cross-entropy loss strategy (figure \ref{fig:6}), 3 features yield the best performance.

\vspace{2 mm}
The Logistic Regression model trained with all the features was compared with the model trained with the best subset of feature from the FS techniques tested. The performance comparison was based on accuracy, specificity, sensitivity (recall), precision, F1-score, and Matthews correlation coefficient (MCC) scores\cite{doc:61b768d98f08073661ec2e97}. These scores were averaged after the k-fold cross-validation. As for the accuracy previously, in this study the performance metrics are intended as individuals classified by the majority of the 26 records classified for each individual.

\vspace{2 mm}
The specificity measures the true negative rate out of the actual negative class, the sensitivity (recall) measures the true positive rate out of the actual positive class, the precision measures the true positive rate out of the predicted positive class, the F1-score  is the harmonic mean of the precision and recall.

\vspace{2 mm}
The MCC score\cite{doc:61cc7baa8f0808703bd4a06e} measures the quality of a binary classifier, a correlation coefficient between predicted and actual class. It ranges between -1, the worst classifier and +1, the best classifier, and a value of zero indicates a random classifier (see Eq.~\ref{eq:mcc}).
\begin{equation} \label{eq:mcc}
  MCC = \frac{TP \times TN - FP \times FN}
                  {\sqrt{(TP + FP)(TP + FN)(TN + FP)(TN + FN)}}
\end{equation}

\vspace{2 mm}
\section{Experimental Results}
The results are summarized in the Table \ref{tab:results summary}. The classes in the dataset are balanced, hence there is no difference between Accuracy and Sensitivity. In the biomedical field the Sensitivity is an important metric of how correctly a classifier detect individuals with a condition, hence it’s possible to early diagnosis a disease and increases the chance of therapies treatment that might help the symptoms of a disease from becoming worse. All the FS technique lead to better performance than the model trained with all features.

The SFS technique with MIN cross-entropy loss strategy is the technique that achieve the best performance, with a subset of 3 features, Jitter (ppq5), Shimmer (apq11) and the correlation between NHR and HNR.

The ANOVA and SFS technique both with MAX accuracy strategy achieve the same performance and yield to the same subset of 1 feature, the Jitter (ppq5). The ANOVA technique with MIN cross-entropy loss strategy and LASSO technique with both MAX accuracy strategy and MIN cross-entropy loss strategy achieve the lowest performance and yields a subset of 7, 4 and 9 feature respectively. They all include the features Jitter (ppq5) and Shimmer (apq11).

\section{Conclusions}
From the test performed with the FS techniques, by reducing the number of feature to train a model, it’s possible to speed up ML inference and increase at same time the model performance for diagnosis Parkinson’s disease. The approach proposed is only part of all the possible classifiers and feature techniques that can be applied but it gives the idea of the opportunities in this field of research. Regarding the features, Jitter (ppq5), Shimmer (apq11) are common features selected by each FS technique. Next step is to understand why and how these subset of features yield the most information that lead to improving the classifier performance for PD detection.

\begin{acks}
This research is the output of the course COMP-592DL Project in Data Science attended during the Master of Science in Data Science at the University of Nicosia.

Acknowledge and thanks to Dr Ioannis Katakis and Dr Kalia Orphanou, for the help and the support, not only during the writing of this paper but also during the entire course of study.
\end{acks}

\bibliographystyle{ACM-Reference-Format}
\bibliography{sample-base}


\begin{thebibliography}{11}


\ifx \showCODEN    \undefined \def \showCODEN     #1{\unskip}     \fi
\ifx \showDOI      \undefined \def \showDOI       #1{#1}\fi
\ifx \showISBNx    \undefined \def \showISBNx     #1{\unskip}     \fi
\ifx \showISBNxiii \undefined \def \showISBNxiii  #1{\unskip}     \fi
\ifx \showISSN     \undefined \def \showISSN      #1{\unskip}     \fi
\ifx \showLCCN     \undefined \def \showLCCN      #1{\unskip}     \fi
\ifx \shownote     \undefined \def \shownote      #1{#1}          \fi
\ifx \showarticletitle \undefined \def \showarticletitle #1{#1}   \fi
\ifx \showURL      \undefined \def \showURL       {\relax}        \fi
\providecommand\bibfield[2]{#2}
\providecommand\bibinfo[2]{#2}
\providecommand\natexlab[1]{#1}
\providecommand\showeprint[2][]{arXiv:#2}

\bibitem[Chandrashekar and Sahin(2014)]%
        {doc:61c9e5a38f08dfade47e620e}
\bibfield{author}{\bibinfo{person}{Girish Chandrashekar} {and}
  \bibinfo{person}{Ferat Sahin}.} \bibinfo{year}{2014}\natexlab{}.
\newblock \showarticletitle{A survey on feature selection methods}.
\newblock \bibinfo{journal}{\emph{Computers \& Electrical Engineering}}
  \bibinfo{volume}{40}, \bibinfo{number}{1} (\bibinfo{year}{2014}),
  \bibinfo{pages}{16--28}.
\newblock


\bibitem[Foster et~al\mbox{.}(2014)]%
        {doc:61c9e1b88f083801efa8c721}
\bibfield{author}{\bibinfo{person}{Kenneth~R. Foster}, \bibinfo{person}{Robert
  Koprowski}, {and} \bibinfo{person}{Joseph~D. Skufca}.}
  \bibinfo{year}{2014}\natexlab{}.
\newblock \showarticletitle{Machine learning, medical diagnosis, and biomedical
  engineering research-commentary}.
\newblock \bibinfo{journal}{\emph{Biomedical engineering online}}
  \bibinfo{volume}{13}, \bibinfo{number}{1} (\bibinfo{year}{2014}),
  \bibinfo{pages}{1--9}.
\newblock


\bibitem[Jankovic(2008)]%
        {doc:61c9e0678f080835edec03ca}
\bibfield{author}{\bibinfo{person}{Joseph Jankovic}.}
  \bibinfo{year}{2008}\natexlab{}.
\newblock \showarticletitle{Parkinson’s disease: clinical features and
  diagnosis}.
\newblock \bibinfo{journal}{\emph{Journal of neurology, neurosurgery \&
  psychiatry}} \bibinfo{volume}{79}, \bibinfo{number}{4}
  (\bibinfo{year}{2008}), \bibinfo{pages}{368--376}.
\newblock
\newblock
\shownote{pmid:18344392}.


\bibitem[Kononenko et~al\mbox{.}(1997)]%
        {doc:61c9e1ce8f0808703bd4405b}
\bibfield{author}{\bibinfo{person}{Igor Kononenko}, \bibinfo{person}{Ivan
  Bratko}, {and} \bibinfo{person}{Matjaž Kukar}.}
  \bibinfo{year}{1997}\natexlab{}.
\newblock \showarticletitle{Application of machine learning to medical
  diagnosis}.
\newblock \bibinfo{journal}{\emph{Machine Learning and Data Mining: Methods and
  Applications}}  \bibinfo{volume}{389} (\bibinfo{year}{1997}),
  \bibinfo{pages}{408}.
\newblock


\bibitem[Little et~al\mbox{.}(2008)]%
        {doc:61c9dfe68f08bf9aa01081f7}
\bibfield{author}{\bibinfo{person}{Max Little}, \bibinfo{person}{Patrick
  McSharry}, \bibinfo{person}{Eric Hunter}, \bibinfo{person}{Jennifer
  Spielman}, {and} \bibinfo{person}{Lorraine Ramig}.}
  \bibinfo{year}{2008}\natexlab{}.
\newblock \showarticletitle{Suitability of dysphonia measurements for
  telemonitoring of Parkinson’s disease}.
\newblock \bibinfo{journal}{\emph{Nature Precedings}} (\bibinfo{year}{2008}),
  \bibinfo{pages}{1}.
\newblock


\bibitem[Matthews(1975)]%
        {doc:61cc7baa8f0808703bd4a06e}
\bibfield{author}{\bibinfo{person}{Brian~W. Matthews}.}
  \bibinfo{year}{1975}\natexlab{}.
\newblock \showarticletitle{Comparison of the predicted and observed secondary
  structure of T4 phage lysozyme}.
\newblock \bibinfo{journal}{\emph{Biochimica et Biophysica Acta (BBA)-Protein
  Structure}} \bibinfo{volume}{405}, \bibinfo{number}{2}
  (\bibinfo{year}{1975}), \bibinfo{pages}{442--451}.
\newblock


\bibitem[Ng(2004)]%
        {doc:61c9fba88f0808703bd441b5}
\bibfield{author}{\bibinfo{person}{Andrew~Y. Ng}.}
  \bibinfo{year}{2004}\natexlab{}.
\newblock \showarticletitle{Feature selection, L 1 vs. L 2 regularization, and
  rotational invariance}. In \bibinfo{booktitle}{\emph{Proceedings of the
  twenty-first international conference on Machine learning}}.
  \bibinfo{pages}{78}.
\newblock


\bibitem[Sakar et~al\mbox{.}(2013)]%
        {doc:61b768d98f08073661ec2e97}
\bibfield{author}{\bibinfo{person}{B.~E. Sakar}, \bibinfo{person}{M.~E.
  Isenkul}, \bibinfo{person}{C.~O. Sakar}, \bibinfo{person}{A. Sertbas},
  \bibinfo{person}{F. Gurgen}, \bibinfo{person}{S. Delil}, \bibinfo{person}{H.
  Apaydin}, {and} \bibinfo{person}{O. Kursun}.}
  \bibinfo{year}{2013}\natexlab{}.
\newblock \bibinfo{title}{Collection and Analysis of a Parkinson Speech Dataset
  With Multiple Types of Sound Recordings}.
\newblock , \bibinfo{numpages}{828-834}~pages.
\newblock
\showISBNx{2168-2208}
\urldef\tempurl%
\url{https://doi.org/10.1109/JBHI.2013.2245674}
\showDOI{\tempurl}
\newblock
\shownote{ID: 1}.


\bibitem[Sakar and Kursun(2010)]%
        {RefWorks:RefID:49-sakar2010telediagnosis}
\bibfield{author}{\bibinfo{person}{C.~Okan Sakar} {and} \bibinfo{person}{Olcay
  Kursun}.} \bibinfo{year}{2010}\natexlab{}.
\newblock \showarticletitle{Telediagnosis of Parkinson’s disease using
  measurements of dysphonia}.
\newblock \bibinfo{journal}{\emph{Journal of medical systems}}
  \bibinfo{volume}{34}, \bibinfo{number}{4} (\bibinfo{year}{2010}),
  \bibinfo{pages}{591--599}.
\newblock


\bibitem[Singh et~al\mbox{.}(2007)]%
        {doc:61cc1d7c8f08bf9aa010cb4d}
\bibfield{author}{\bibinfo{person}{Neha Singh}, \bibinfo{person}{Viness
  Pillay}, {and} \bibinfo{person}{Yahya~E. Choonara}.}
  \bibinfo{year}{2007}\natexlab{}.
\newblock \showarticletitle{Advances in the treatment of Parkinson's disease}.
\newblock \bibinfo{journal}{\emph{Progress in neurobiology}}
  \bibinfo{volume}{81}, \bibinfo{number}{1} (\bibinfo{year}{2007}),
  \bibinfo{pages}{29--44}.
\newblock


\bibitem[Yeo and Johnson(2000)]%
        {doc:61cc4e8e8f0808703bd49825}
\bibfield{author}{\bibinfo{person}{In‐Kwon Yeo} {and}
  \bibinfo{person}{Richard~A. Johnson}.} \bibinfo{year}{2000}\natexlab{}.
\newblock \showarticletitle{A new family of power transformations to improve
  normality or symmetry}.
\newblock \bibinfo{journal}{\emph{Biometrika}} \bibinfo{volume}{87},
  \bibinfo{number}{4} (\bibinfo{year}{2000}), \bibinfo{pages}{954--959}.
\newblock


\end{thebibliography}



\end{document}